\documentclass{article}

\usepackage{microtype}
\usepackage{array}
\usepackage{graphicx}
\usepackage{subfigure}
\usepackage{silence}
\usepackage{multirow}
\usepackage{amsmath}
\usepackage{amsfonts}
\usepackage{tabularx}
\usepackage{float}
\usepackage{booktabs} 

\newcolumntype{x}[1]{>{\centering\arraybackslash}p{#1}}
\newcolumntype{Y}{>{\centering\arraybackslash}X}

\usepackage{hyperref}

\usepackage[accepted]{icml2019}

\icmltitlerunning{Graph U-Nets}

\begin{document}

\twocolumn[
\icmltitle{Graph U-Nets}



\icmlsetsymbol{equal}{*}

\begin{icmlauthorlist}
\icmlauthor{Hongyang Gao}{to}
\icmlauthor{Shuiwang Ji}{to}
\end{icmlauthorlist}

\icmlaffiliation{to}{Department of Computer Science \& Engineering, Texas A\&M University, TX, USA}
\icmlcorrespondingauthor{Hongyang Gao}{hongyang.gao@tamu.edu}
\icmlcorrespondingauthor{Shuiwang Ji}{sji@tamu.edu}

\icmlkeywords{Machine Learning, ICML}

\vskip 0.3in
]



\printAffiliationsAndNotice{}  

\begin{abstract}
We consider the problem of representation learning for graph
data. Convolutional neural networks can naturally operate on images, but have
significant challenges in dealing with graph data. Given images are special
cases of graphs with nodes lie on 2D lattices, graph embedding tasks have a
natural correspondence with image pixel-wise prediction tasks such as
segmentation. While encoder-decoder architectures like U-Nets have been
successfully applied on many image pixel-wise prediction tasks, similar
methods are lacking for graph data. This is due to the fact that pooling and
up-sampling operations are not natural on graph data. To address these
challenges, we propose novel graph pooling~(gPool) and unpooling~(gUnpool)
operations in this work. The gPool layer adaptively selects some nodes to form
a smaller graph based on their scalar projection values on a trainable
projection vector. We further propose the gUnpool layer as the inverse
operation of the gPool layer. The gUnpool layer restores the graph into its
original structure using the position information of nodes selected in the
corresponding gPool layer. Based on our proposed gPool and gUnpool layers, we
develop an encoder-decoder model on graph, known as the graph U-Nets. Our
experimental results on node classification and graph
classification tasks demonstrate that our methods achieve consistently better
performance than previous models.
\end{abstract}

\section{Introduction}

Convolutional neural networks~(CNNs)~\citep{lecun2012efficient} have
demonstrated great capability in various challenging artificial
intelligence tasks, especially in fields of computer
vision~\citep{he2017mask,huang2016densely} and natural language
processing~\citep{bahdanau2014neural}. One
common property behind these tasks is that both images and texts
have grid-like structures. Elements on feature maps have locality
and order information, which enables the application of
convolutional operations~\citep{defferrard2016convolutional}.

In practice, many real-world data can be naturally represented as graphs such
as social and biological networks. Due to the great success of CNNs on
grid-like data, applying them on graph data~\citep{gori2005new,scarselli2009graph}
is particularly appealing. Recently, there have been many attempts to extend
convolutions to graph data~(GNNs)
\citep{kipf2016semi,velivckovic2017graph,gao2018large}. One common use of
convolutions on graphs is to compute node
representations~\citep{hamilton2017inductive,ying2018hierarchical}. With
learned node representations, we can perform various tasks on graphs such as
node classification and link prediction.

Images can be considered as special cases of graphs, in which nodes
lie on regular 2D lattices. It is this special structure that
enables the use of convolution and pooling operations on images.
Based on this relationship, node classification and embedding tasks
have a natural correspondence with pixel-wise prediction tasks such
as image segmentation~\citep{noh2015learning,gao2017efficient,jegou2017one}. In
particular, both tasks aim to make predictions for each input unit,
corresponding to a pixel on images or a node in graphs.
In the computer vision field, pixel-wise prediction tasks have
achieved major advances recently. Encoder-decoder architectures like
the U-Net~\citep{ronneberger2015u} are state-of-the-art methods for
these tasks. It is thus highly interesting to develop U-Net-like
architectures for graph data. In addition to convolutions, pooling
and up-sampling operations are essential building blocks in these
architectures. However, extending these operations to graph data is
highly challenging. Unlike grid-like data such as images and texts,
nodes in graphs have no spatial locality and order information as
required by regular pooling operations.

To bridge the above gap, we propose novel graph pooling~(gPool) and
unpooling~(gUnpool) operations in this work. Based on these two
operations, we propose U-Net-like architectures for graph data. The
gPool operation samples some nodes to form a smaller graph based on
their scalar projection values on a trainable projection vector. As
an inverse operation of gPool, we propose a corresponding graph
unpooling~(gUnpool) operation, which restores the graph to its
original structure with the help of locations of nodes selected in
the corresponding gPool layer. Based on the gPool and gUnpool
layers, we develop graph U-Nets, which allow high-level feature
encoding and decoding for network embedding. Experimental results on node
classification and graph classification tasks
demonstrate the effectiveness of our proposed methods as compared to
previous methods.

\section{Related Work}

Recently, there has been a rich line of research on graph neural
networks~\citep{gilmer2017neural}. Inspired by the first order graph
Laplacian methods, \cite{kipf2016semi} proposed graph convolutional
networks~(GCNs), which achieved promising performance on graph node
classification tasks. The layer-wise forward-propagation operation
of GCNs is defined as:
\begin{equation}
\begin{aligned}
  X_{\ell+1} =
  \sigma(\hat{D}^{-\frac{1}{2}}\hat{A}\hat{D}^{-\frac{1}{2}}X_{\ell}W_{\ell}),
\end{aligned}\label{eq:gcn}
\end{equation}
where $\hat A = A + I$ is used to add self-loops in the input
adjacency matrix $A$, $X_{\ell}$ is the feature matrix of layer
$\ell$. The GCN layer uses the diagonal node degree matrix $\hat{D}$
to normalize $\hat{A}$. $W_{\ell}$ is a trainable weight matrix that
applies a linear transformation to feature vectors. GCNs essentially
perform aggregation and transformation on node features without
learning trainable filters. \cite{hamilton2017inductive} tried to
sample a fixed number of neighboring nodes to keep the computational
footprint consistent. \cite{velivckovic2017graph} proposed to use
attention mechanisms to enable different weights for neighboring
nodes. \cite{schlichtkrull2018modeling} used relational graph
convolutional networks for link prediction and entity
classification. Some studies applied GNNs to graph classification
tasks~\citep{duvenaud2015convolutional,dai2016discriminative,zhang2018end}.
\cite{bronstein2017geometric} discussed possible
ways of applying deep learning on graph data. \cite{henaff2015deep}
and \cite{bruna2014spectral} proposed to use spectral networks for
large-scale graph classification tasks. Some studies also applied
graph kernels on traditional computer vision
tasks~\citep{gama2019convolutional,fey2018splinecnn,monti2017geometric}.

In addition to convolution, some studies tried to extend pooling operations to
graphs. \cite{defferrard2016convolutional} proposed to use binary tree
indexing for graph coarsening, which fixes indices of nodes before applying
1-D pooling operations. \cite{simonovsky2017dynamic} used deterministic graph
clustering algorithm to determine pooling patterns.
\cite{ying2018hierarchical} used an assignment matrix to achieve pooling by
assigning nodes to different clusters of the next layer.


\section{Graph U-Nets}

In this section, we introduce the graph pooling~(gPool) layer and
graph unpooling~(gUnpool) layer. Based on these two new layers, we
develop the graph U-Nets for node classification tasks.

\subsection{Graph Pooling Layer}

\begin{figure*}[t] \includegraphics[width=\textwidth]{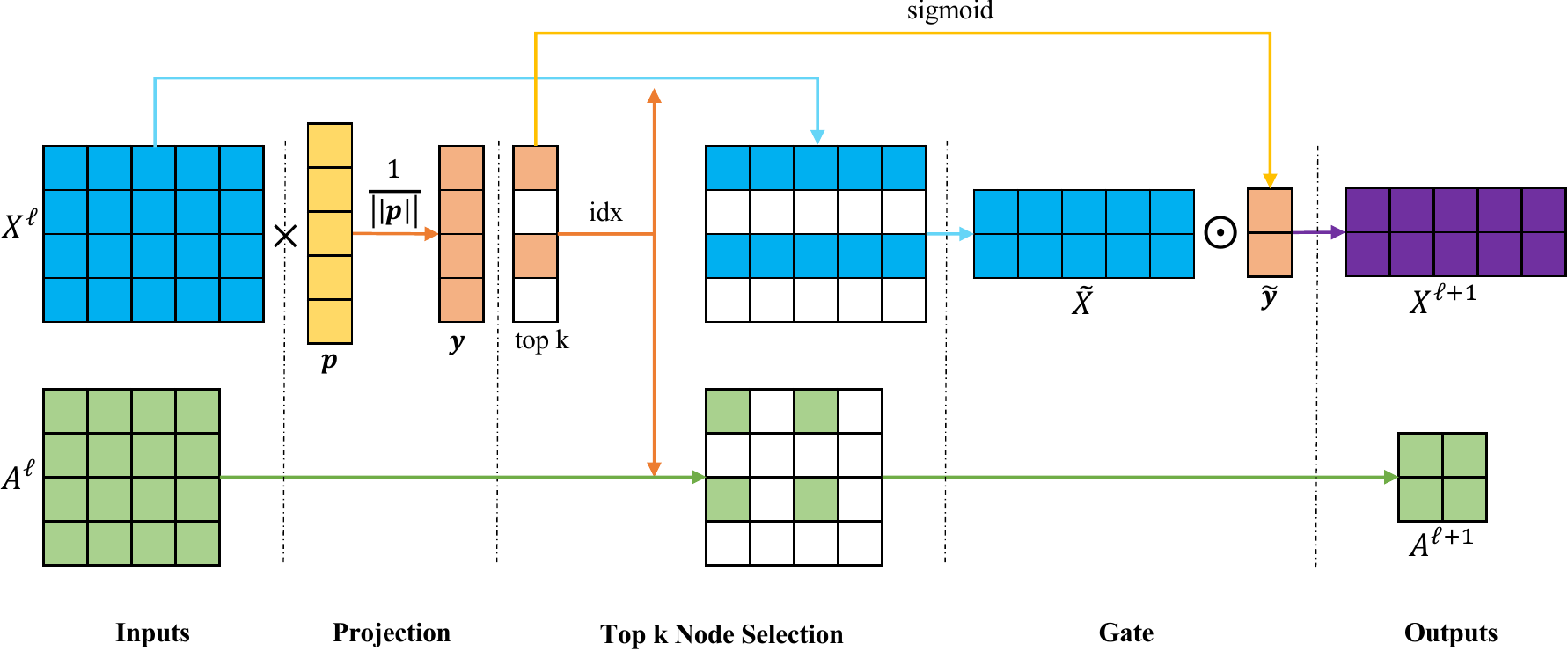}
\caption{An illustration of the proposed graph pooling layer with
$k=2$. $\times$ and $\odot$ denote matrix multiplication and
element-wise product, respectively. We consider a graph with 4
nodes, and each node has 5 features. By processing this graph, we
obtain the adjacency matrix $A^\ell \in \mathbb{R}^{4 \times 4}$ and
the input feature matrix $X^\ell \in \mathbb{R}^{4 \times 5} $ of
layer $\ell$. In the projection stage, $\mathbf p \in \mathbb{R}^{5}
$ is a trainable projection vector. By matrix multiplication and
$\mbox{sigmoid}(\cdot)$, we obtain $\mathbf y$ that are scores estimating
scalar projection values of each node to the projection vector.
By using $k=2$, we
select two nodes with the highest scores and record their indices in
the top-k-node selection stage. We use the indices to extract the
corresponding nodes to form a new graph, resulting in the pooled
feature map $\tilde X^{\ell}$ and new corresponding adjacency matrix
$A^{\ell+1}$. At the gate stage, we perform element-wise
multiplication between $\tilde X^{\ell}$ and the selected node
scores vector $\mathbf{\tilde y}$, resulting in $X^{\ell+1}$. This
graph pooling layer outputs $A^{\ell+1}$ and $X^{\ell+1}$.} \label{fig:gpool}
\end{figure*}

Pooling layers play important roles in CNNs on grid-like data. They
can reduce sizes of feature maps and enlarge receptive fields,
thereby giving rise to better generalization and
performance~\citep{yu2015multi}. On grid-like data such as images,
feature maps are partitioned into non-overlapping rectangles, on
which non-linear down-sampling functions like maximum are applied.
In addition to local pooling, global pooling
layers~\citep{zhao2015self} perform down-sampling operations on all
input units, thereby reducing each feature map to a single number.
In contrast, $k$-max pooling layers~\citep{blunsom2014convolutional}
select the $k$-largest units out of each feature map.

However, we cannot directly apply these pooling operations to
graphs. In particular, there is no locality information among nodes
in graphs. Thus the partition operation is not applicable on graphs. The
global pooling operation will reduce all nodes to one single node, which
restricts the flexibility of networks. The $k$-max pooling operation
outputs the $k$-largest units that may come from different nodes in graphs,
resulting in inconsistency in the connectivity of selected nodes.

In this section, we propose the graph pooling~(gPool) layer to
enable down-sampling on graph data. In this layer, we adaptively
select a subset of nodes to form a new but smaller graph. To this
end, we employ a trainable projection vector $\mathbf{p}$. By
projecting all node features to 1D, we can perform $k$-max pooling
for node selection. Since the selection is based on 1D footprint of
each node, the connectivity in the new graph is consistent across
nodes. Given a node $i$ with its feature vector $\mathbf{x}_i$, the
scalar projection of $\mathbf{x}_i$ on $\mathbf{p}$ is $y_i =
\mathbf{x_i} \mathbf{p} / \lVert \mathbf{p} \rVert$. Here, $y_i$
measures how much information of node $i$ can be retained when
projected onto the direction of $\mathbf{p}$. By sampling nodes, we
wish to preserve as much information as possible from the original
graph. To achieve this, we select nodes with the largest scalar
projection values on $\mathbf{p}$ to form a new graph.

Suppose there are $N$ nodes in a graph $\mathbb G$ and each of which
contains $C$ features. The graph can be represented by two matrices;
those are the adjacency matrix $A^{\ell} \in \mathbb{R}^{N \times N}
$ and the feature matrix $X^{\ell} \in \mathbb{R}^{N\times C}$. 
Each non-zero entry in the adjacency matrix $A$ represents an edge between two nodes in the graph.
Each row
vector $\mathbf{x}^{\ell}_i$ in the feature matrix $X^{\ell}$ denotes the feature
vector of node $i$ in the graph. The layer-wise propagation rule of the
graph pooling layer $\ell$ is defined as:
\begin{equation}
\begin{aligned}
  \mathbf y & = X^{\ell} \mathbf p^{\ell} / \lVert \mathbf p^{\ell} \rVert, \\
  \mbox{idx} &= \mbox{rank}(\mathbf y, k), \\
 \tilde{\mathbf y} &= \mbox{sigmoid} (\mathbf y(\mbox{idx})), \\
  \tilde X^{\ell} & = X^{\ell}(\mbox{idx}, :), \\
  A^{\ell+1} &= A^{\ell}(\mbox{idx}, \mbox{idx}), \\
  X^{\ell+1} &= \tilde X^{\ell} \odot \left(\tilde{\mathbf y} \mathbf{1}_C^{T}\right),
\end{aligned}\label{eq:gpool}
\end{equation}
where $k$ is the number of nodes selected in the new graph.
$\mbox{rank}(\mathbf y, k)$ is the operation of node ranking, which
returns indices of the $k$-largest values in $\mathbf y$. The
$\mbox{idx}$ returned by $\mbox{rank}(\mathbf y, k)$ contains the
indices of nodes selected for the new graph. $A^{\ell}(\mbox{idx},
\mbox{idx})$ and $X^{\ell}(\mbox{idx}, :)$ perform the row and/or
column extraction to form the adjacency matrix and the feature
matrix for the new graph. $\mathbf y(\mbox{idx})$ extracts values in
$\mathbf y$ with indices idx followed by a $\mbox{sigmoid}$ operation.
$\mathbf 1_{C}\in \mathbb{R}^{C}$ is a vector of size $C$ with all
components being 1, and $\odot$ represents the element-wise matrix
multiplication.

$X^{\ell}$ is the feature matrix with row vectors $\mathbf
x^{\ell}_1, \mathbf x^{\ell}_2, \cdots, \mathbf x^{\ell}_N$, each of
which corresponds to a node in the graph. We first compute the
scalar projection of $X^{\ell}$ on $\mathbf{p}^{\ell}$, resulting in
$\mathbf y = [ y_1, y_2, \cdots, y_N ]^T$ with each $y_i$ measuring
the scalar projection value of each node on the projection vector
$\mathbf p^{\ell}$. Based on the scalar projection vector $\mathbf
y$, $\mbox{rank}(\cdot)$ operation ranks values and returns the
$k$-largest values in $\mathbf y$. Suppose the $k$-selected indices
are $i_1, i_2, \cdots, i_k $ with $i_m < i_n$ and $1 \le m < n \le
k$. Note that the index selection process preserves the position
order information in the original graph. With indices idx, we
extract the adjacency matrix $A^{\ell} \in \mathbb{R}^{k\times k}$
and the feature matrix $\tilde X^{\ell} \in \mathbb{R}^{k \times C}$
for the new graph. Finally, we employ a gate operation to control
information flow. With selected indices idx, we obtain the gate
vector $\tilde y \in \mathbb{R}^k$ by applying $\mbox{sigmoid}$ to each
element in the extracted scalar projection vector. Using
element-wise matrix product of $\tilde X^{\ell}$ and $\mathbf{\tilde
y} \mathbf 1^T_C$, information of selected nodes is controlled. The
$i$th row vector in $X^{\ell+1}$ is the product of the $i$th row
vector in $X^{\ell}$ and the $i$th scalar value in $\tilde y$.

Notably, the gate operation makes the projection vector $\mathbf p$
trainable by back-propagation~\citep{lecun2012efficient}. Without the
gate operation, the projection vector $\mathbf p$ produces discrete
outputs, which makes it not trainable by back-propagation.
Figure~\ref{fig:gpool} provides an illustration of our proposed
graph pooling layer. Compared to pooling operations used in
grid-like data, our graph pooling layer employs extra training
parameters in projection vector $\mathbf p$. We will show that the
extra parameters are negligible but can boost performance.

\subsection{Graph Unpooling Layer}

\begin{figure*}[t]  \centering
\includegraphics[width=0.8\textwidth]{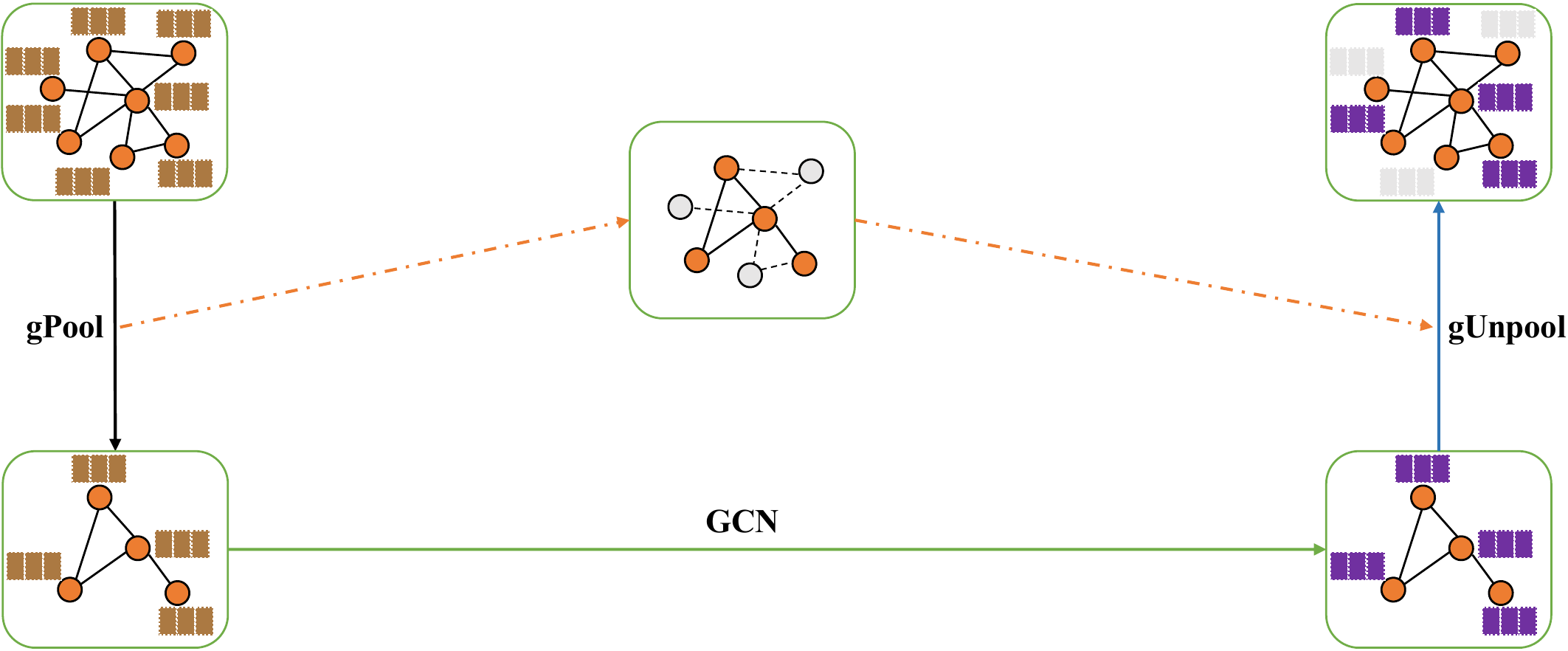}
\caption{An illustration of the proposed graph unpooling~(gUnpool)
layer. In this example, a graph with 7 nodes is down-sampled using a
gPool layer, resulting in a coarsened graph with 4 nodes and
position information of selected nodes. The corresponding gUnpool
layer uses the position information to reconstruct the original
graph structure by using empty feature vectors for unselected
nodes.} \label{fig:gunpool} \end{figure*}

Up-sampling operations are important for encoder-decoder networks
such as U-Net. The encoders of networks usually employ pooling
operations to reduce feature map size and increase receptive field.
While in decoders, feature maps need to be up-sampled to restore
their original resolutions. On grid-like data like images, there are
several up-sampling operations such as the
deconvolution~\citep{isola2017image,Zhao2015StackedWA} and
unpooling layers~\citep{long2015fully}.
However, such operations are not currently
available on graph data.

To enable up-sampling operations on graph data, we propose the graph
unpooling~(gUnpool) layer, which performs the inverse operation of
the gPool layer and restores the graph into its original structure.
To achieve this, we record the locations of nodes selected in the
corresponding gPool layer and use this information to place nodes
back to their original positions in the graph. Formally, we propose
the layer-wise propagation rule of graph unpooling layer as
\begin{equation}
\begin{aligned}
  X^{\ell+1} &= \mbox{distribute}(0_{N\times C}, X^{\ell}, \mbox{idx}), \\
\end{aligned}\label{eq:gunpool}
\end{equation}
where $\mbox{idx} \in \mathbb{Z}^{*k}$ contains indices of selected
nodes in the corresponding gPool layer that reduces the graph size
from $N$ nodes to $k$ nodes. $X^{\ell} \in \mathbb{R}^{k \times C}$
are the feature matrix of the current graph, and $0_{N\times C}$ are
the initially empty feature matrix for the new graph.
$\mbox{distribute}(0_{N\times C}, X^{\ell}, \mbox{idx})$ is the
operation that distributes row vectors in $X^{\ell}$ into
$0_{N\times C}$ feature matrix according to their corresponding
indices stored in $\mbox{idx}$. In $X^{\ell+1}$, row vectors with
indices in $\mbox{idx}$ are updated by row vectors in $X^{\ell}$,
while other row vectors remain zero.

\subsection{Graph U-Nets Architecture}\label{sec:gunet}

\begin{figure*}[t] \includegraphics[width=\textwidth]{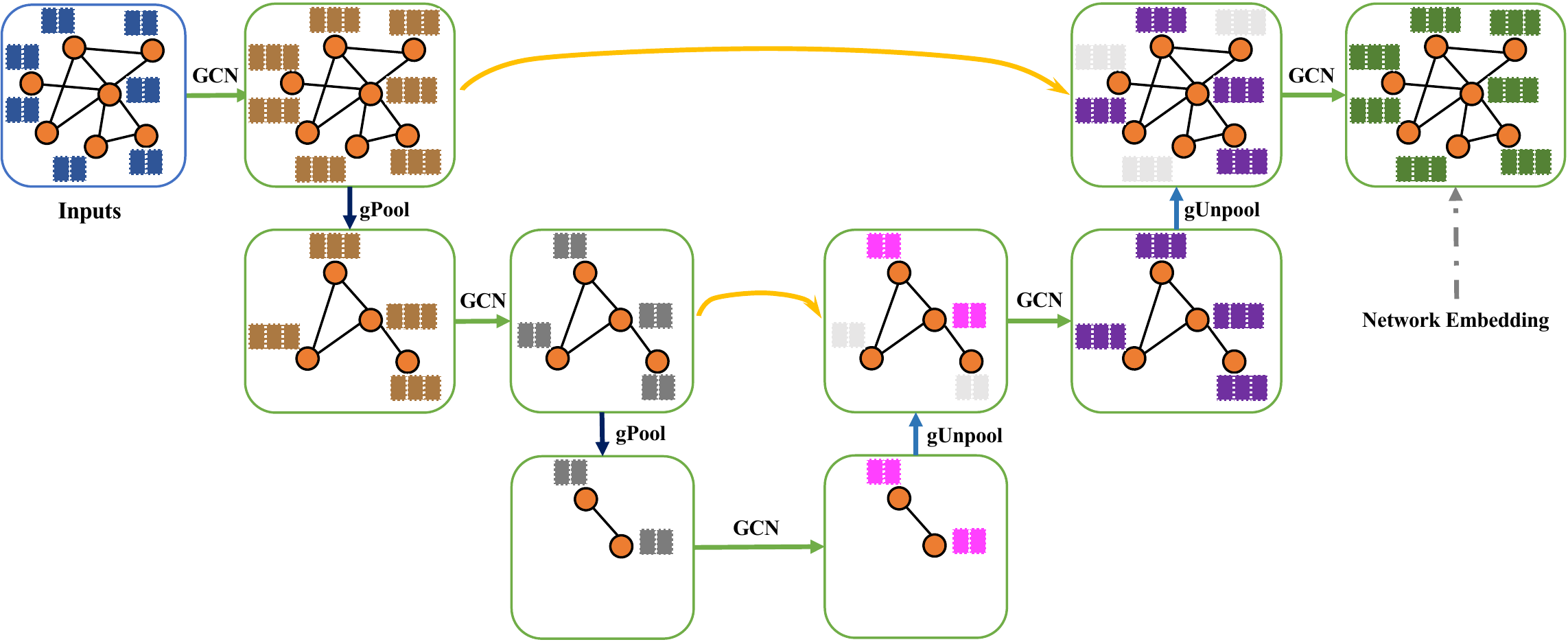}
\caption{An illustration of the proposed graph U-Nets~(g-U-Nets). In
this example, each node in the input graph has two features. The
input feature vectors are transformed into low-dimensional
representations using a GCN layer. After that, we stack two encoder
blocks, each of which contains a gPool layer and a GCN layer. In the
decoder part, there are also two decoder blocks. Each block consists
of a gUnpool layer and a GCN layer. For blocks in the same level,
encoder block uses skip connection to fuse the low-level spatial
features from the encoder block. The output feature vectors of nodes
in the last layer are network embedding, which can be used for
various tasks such as node classification and link prediction. }
\label{fig:unet}
\end{figure*}

It is well-known that encoder-decoder networks like U-Net achieve
promising performance on pixel-wise prediction tasks, since they can
encode and decode high-level features while maintaining local
spatial information. Similar to pixel-wise prediction
tasks~\citep{gong2013deep,ronneberger2015u}, node classification
tasks aim to make a prediction for each input unit. Based on our
proposed gPool and gUnpool layers, we propose our graph
U-Nets~(g-U-Nets) architecture for node classification tasks.

In our graph U-Nets~(g-U-Nets), we first apply a graph embedding
layer to convert nodes into low-dimensional representations, since
original inputs of some dataset like Cora~\citep{sen2008collective}
usually have very high-dimensional feature vectors. After the graph
embedding layer, we build the encoder by stacking several encoding
blocks, each of which contains a gPool layer followed by a GCN
layer. gPool layers reduce the size of graph to encode higher-order
features, while GCN layers are responsible for aggregating
information from each node's first-order information. In the decoder
part, we stack the same number of decoding blocks as in the encoder
part. Each decoder block is composed of a gUnpool layer and a GCN
layer. The gUnpool layer restores the graph into its higher
resolution structure, and the GCN layer aggregates information from
the neighborhood. There are skip-connections between corresponding
blocks of encoder and decoder layers, which transmit spatial
information to decoders for better performance. The skip-connection
can be either feature map addition or concatenation. Finally, we
employ a GCN layer for final predictions before the softmax
function. Figure~\ref{fig:unet} provides an illustration of a sample
g-U-Nets with two blocks in encoder and decoder.
Notably, there is a GCN layer before each gPool layer, thereby
enabling gPool layers to capture the topological information in
graphs implicitly.

\subsection{Graph Connectivity Augmentation via Graph Power}\label{sec:aug}

In our proposed gPool layer, we sample some important nodes to form
a new graph for high-level feature encoding. Since related edges are
removed when removing nodes in gPool, the nodes in the pooled graph
might become isolated. This may influence the information
propagation in subsequent layers, especially when GCN layers are
used to aggregate information from neighboring nodes. We need to
increase connectivity among nodes in the pooled graph. To address
this problem, we propose to use the $k^{th}$ graph power
$\mathbb{G}^k$ to increase the graph connectivity. This operation
builds links between nodes whose distances are at most $k$
hops~\citep{chepuri2016subsampling}. In this work, we employ $k=2$
since there is a GCN layer before each gPool layer to aggregate
information from its first-order neighboring nodes. Formally, we
replace the fifth equation in Eq~\ref{eq:gpool} by:
\begin{equation}
A^{2} = A^{\ell}A^{\ell}, \,\,\,\,\,\,\,\,A^{\ell+1} =
A^{2}(\mbox{idx}, \mbox{idx}), \label{eq:gaug}
\end{equation}
where $A^2 \in \mathbb{R}^{N \times N}$ is the $2^{nd}$ graph power.
Now, the graph sampling is performed on the augmented graph with
better connectivity.
%

\subsection{Improved GCN Layer}

In Eq.~\ref{eq:gcn}, the adjacency matrix before normalization is
computed as $\hat A = A + I$ in which a self-loop is added to each
node in the graph. When performing information aggregation, the same
weight is given to node's own feature vector and its neighboring
nodes. In this work, we wish to give a higher weight to node's own
feature vector, since its own feature should be more important for
prediction. To this end, we change the calculation to $\hat{A} = A +
2I$ by imposing larger weights on self loops in the graph, which is
common in graph processing. All experiments in this work use this
modified version of GCN layer for better performance.

\section{Experimental Study}

In this section, we evaluate our gPool and gUnpool layers based on
the g-U-Nets proposed in Section~\ref{sec:gunet}. We compare our
networks with previous state-of-the-art models on node
classification and graph classification tasks.
Experimental results show that our methods achieve new
state-of-the-art results in terms of node classification accuracy
and graph classification accuracy. Some ablation
studies are performed to examine the contributions of the proposed
gPool layer, gUnpool layer, and graph connectivity augmentation to
performance improvements. We conduct studies on the relationship
between network depth and node classification performance. We
investigate if additional parameters involved in gPool layers can
increase the risk of over-fitting.

\begin{table*}[t]
\centering \caption{Summary of datasets used in our node
classification
experiments~\citep{yang2016revisiting,zitnik2017predicting}. The
Cora, Citeseer, and Pubmed datasets are used for transductive
learning experiments.} \label{table:transdatasets}
\begin{tabularx}{\textwidth}{  X YYYYYYYY }
    \hline
    \textbf{Dataset} & \textbf{Nodes} &
    \textbf{Features} & \textbf{Classes} & \textbf{Training} &
    \textbf{Validation} & \textbf{Testing} & \textbf{Degree} \\ \hline\hline
    Cora      & 2708   & 1433 & 7   & 140   & 500 & 1000 & 4  \\ \hline
    Citeseer  & 3327   & 3703 & 6   & 120   & 500 & 1000 & 5  \\ \hline
    Pubmed    & 19717  & 500  & 3   & 60    & 500 & 1000 & 6  \\ \hline
    \hline
\end{tabularx}
\end{table*}

\begin{table*}[t]
\centering \caption{Summary of datasets used in our inductive
learning experiments. The D\&D~\citep{dobson2003distinguishing},
PROTEINS~\citep{borgwardt2005protein}, and
COLLAB~\citep{yanardag2015structural} datasets are used for
inductive learning experiments.} \label{table:inducdatasets}
\begin{tabularx}{\textwidth}{  X YYYY }
    \hline
    \textbf{Dataset} & \textbf{Graphs} &
    \textbf{Nodes (max)} & \textbf{Nodes (avg)} & \textbf{Classes} \\ \hline\hline
    D\&D      & 1178   & 5748 & 284.32  & 2  \\ \hline
    PROTEINS  & 1113   & 620  & 39.06   & 2  \\ \hline
    COLLAB    & 5000   & 492  & 74.49   & 3  \\ \hline
    \hline
\end{tabularx}
\end{table*}

\subsection{Datasets}

In experiments, we evaluate our networks on node classification
tasks under transductive learning settings and
graph classification tasks under inductive learning settings.

Under transductive learning settings,
unlabeled data are accessible for training, which enables the
network to learn about the graph structure. To be specific, only
part of nodes are labeled while labels of other nodes in the same
graph remain unknown. We employ three benchmark datasets for this
setting; those are Cora, Citeseer, and Pubmed~\citep{kipf2016semi},
which are summarized in Table~\ref{table:transdatasets}. These datasets
are citation networks, with each node and each edge representing a
document and a citation, respectively. The feature vector of each
node is the bag-of-word representation whose dimension is determined
by the dictionary size. We follow the same experimental settings
in~\citep{kipf2016semi}. For each class, there are 20 nodes for
training, 500 nodes for validation, and 1000 nodes for testing.

Under inductive learning settings, testing data are
not available during training, which means the training process does
not use graph structures of testing data. We evaluate our methods on
relatively large graph datasets selected from common benchmarks used
in graph classification
tasks~\citep{ying2018hierarchical,niepert2016learning,zhang2018end}.
We use protein datasets including
D\&D~\citep{dobson2003distinguishing} and
PROTEINS~\citep{borgwardt2005protein}, the scientific collaboration
dataset COLLAB~\citep{yanardag2015structural}. These data are
summarized in Table~\ref{table:inducdatasets}.

\subsection{Experimental Setup}

We describe the experimental setup for both transductive and inductive
learning settings. For transductive learning tasks, we employ our proposed
g-U-Nets proposed in Section~\ref{sec:gunet}. Since nodes in the three
datasets are associated with high-dimensional features, we employ a GCN layer
to reduce them into low-dimensional representations. In the encoder part, we
stack four blocks, each of which consists of a gPool layer and a GCN layer. We
sample 2000, 1000, 500, 200 nodes in the four gPool layers, respectively.
Correspondingly, the decoder part also contains four blocks. Each decoder
block is composed of a gUnpool layer and a GCN layer. We use addition
operation in skip connections between blocks of encoder and decoder parts.
Finally, we apply a GCN layer for final prediction. For all layers in the
model, we use identity activation function~\cite{gao2018large} after each
GCN layer. To avoid
over-fitting, we apply $L_2$ regularization on weights with $\lambda=0.001$.
Dropout~\citep{srivastava2014dropout} is applied to both adjacency matrices
and feature matrices with keep rates of 0.8 and 0.08, respectively.

For inductive learning tasks, we follow the same
experimental setups in~\cite{zhang2018end} using our g-U-Nets
architecture as described in transductive learning settings for feature
extraction. Since
the sizes of graphs vary in graph classification tasks, we sample
proportions of nodes in four gPool layers; those are 90\%, 70\%,
60\%, and 50\%, respectively. The dropout keep rate imposed on feature
matrices is 0.3.

\begin{table*}[t]
\centering \caption{Results of transductive learning experiments in
terms of node classification accuracies on Cora, Citeseer, and
Pubmed datasets. g-U-Nets denotes our proposed graph U-Nets model.}
\label{table:trans}
\begin{tabularx}{\textwidth}{  lx{3.5cm} YYY  }
    \hline
    \textbf{Models} & \textbf{Cora} & \textbf{Citeseer} & \textbf{Pubmed} \\ \hline\hline
    DeepWalk~\citep{perozzi2014deepwalk}            & 67.2\% & 43.2\%  & 65.3\%   \\ \hline
    Planetoid~\citep{yang2016revisiting}            & 75.7\% & 64.7\%  & 77.2\%   \\ \hline
    Chebyshev~\citep{defferrard2016convolutional}   & 81.2\% & 69.8\%  & 74.4\%   \\ \hline
    GCN~\citep{kipf2016semi}                        & 81.5\% & 70.3\%  & 79.0\%   \\ \hline
    GAT~\citep{velivckovic2017graph}                & 83.0 $\pm$ 0.7\% & 72.5 $\pm$ 0.7\% & 79.0 $\pm$ 0.3\% \\ \hline
    \textbf{g-U-Nets (Ours)}                         & \textbf{84.4 $\pm$ 0.6\%}
                                                    & \textbf{73.2 $\pm$ 0.5\%}
                                                    & \textbf{79.6 $\pm$ 0.2\%} \\ \hline
    \hline
\end{tabularx}
\end{table*}

\begin{table*}[t]
\centering \caption{Results of inductive learning experiments in
terms of graph classification accuracies on D\&D, PROTEINS, and
COLLAB datasets. g-U-Nets denotes our proposed graph U-Nets model.}
\label{table:induc}
\begin{tabularx}{\textwidth}{  lx{3.5cm}  YYY }
    \hline
    \textbf{Models} & \textbf{D\&D} & \textbf{PROTEINS} & \textbf{COLLAB} \\ \hline\hline
    PSCN~\citep{niepert2016learning}                 & 76.27\% & 75.00\%  & 72.60\%   \\ \hline
    DGCNN~\citep{zhang2018end}                       & 79.37\% & 76.26\%  & 73.76\%   \\ \hline
    DiffPool-DET~\citep{ying2018hierarchical}        & 75.47\% & 75.62\%  & \textbf{82.13}\%   \\ \hline
    DiffPool-NOLP~\citep{ying2018hierarchical}       & 79.98\% & 76.22\%  & 75.58\%   \\ \hline
    DiffPool~\citep{ying2018hierarchical}            & 80.64\% & 76.25\%  & 75.48\%   \\ \hline
    \textbf{g-U-Nets (Ours)}                          & \textbf{82.43\%}
                                                     & \textbf{77.68\%}
                                                     & 77.56\% \\ \hline
    \hline
\end{tabularx}
\end{table*}

\subsection{Performance Study}

Under transductive learning settings, we compare our proposed
g-U-Nets with other state-of-the-art models in terms of node
classification accuracy. We report node classification accuracies on
datasets Cora, Citeseer, and Pubmed, and the results are summarized
in Table~\ref{table:trans}. We can observe from the results that our
g-U-Nets achieves consistently better performance than other
networks. For baseline values listed for node classification tasks,
they are the state-of-the-art on these datasets. Our proposed model
is composed of GCN, gPool, and gUnpool layers without involving more
advanced graph convolution layers like GAT. When compared to GCN
directly, our g-U-Nets significantly improves performance on all
three datasets by margins of 2.9\%, 2.9\%, and 0.6\%, respectively.
Note that the only difference between our g-U-Nets and GCN is the
use of encoder-decoder architecture containing gPool and gUnpool
layers. These results demonstrate the effectiveness of g-U-Nets in
network embedding.

Under inductive learning settings, we compared our
methods with other state-of-the-art models on graph classification
tasks with datasets D\&D, PROTEINS, and COLLAB, and the results are
summarized in Table~\ref{table:induc}. We can observe from the
results that our proposed gPool method outperforms
DiffPool~\citep{ying2018hierarchical} by margins of 1.79\% and
1.43\% on the D\&D and PROTEINS datasets. Notably, the result
obtained by DiffPool-DET on COLLAB is significantly higher than all
other methods and the other two DiffPool models. On all three
datasets, our model outperforms baseline models including DiffPool.
In addition, DiffPool claimed that their training utilized auxiliary
task of link prediction to stabilize model performance, which indicates
the instability of DiffPool model. But in our
experiments, we only use graph labels for training without any
auxiliary tasks to stabilize training.

\subsection{Ablation Study of gPool and gUnpool layers}

Although GCNs have been reported to have worse performance when the
network goes deeper~\citep{kipf2016semi}, it may also be argued that
the performance improvement over GCN in Table~\ref{table:trans} is
due to the use of a deeper network architecture. In this section, we
investigate the contributions of gPool and gUnpool layers to the
performance of g-U-Nets. We conduct experiments by removing all
gPool and gUnpool layers from our g-U-Nets, leading to a network
with only GCN layers with skip connections.
Table~\ref{table:gunet_vs_gunet_no_pool} provides the comparison
results between g-U-Nets with and without gPool or gUnpool layers.
The results show that g-U-Nets have better performance over g-U-Nets
without gPool or gUnpool layers by margins of 2.3\%, 1.6\% and 0.5\% on Cora, Citeseer, and Pubmed
datasets, respectively. These results
demonstrate the contributions of gPool and gUnpool layers to
performance improvement. When considering the difference between the
two models in terms of architecture, g-U-Nets enable higher level
feature encoding, thereby resulting in better generalization and
performance.


\begin{table*}[!th]
\centering \caption{Comparison of g-U-Nets with and without gPool or
gUnpool layers in terms of node classification accuracy on Cora,
Citeseer, and Pubmed datasets.} \label{table:gunet_vs_gunet_no_pool}
\begin{tabularx}{\textwidth}{  lx{3cm}   YYY }
    \hline
    \textbf{Models}       & \textbf{Cora} & \textbf{Citeseer} & \textbf{Pubmed} \\ \hline\hline
    g-U-Nets without gPool or gUnpool        & 82.1 $\pm$ 0.6\%   & 71.6 $\pm$ 0.5\% & 79.1 $\pm$ 0.2\% \\ \hline
    \textbf{g-U-Nets (Ours)}                           & \textbf{84.4 $\pm$ 0.6\%}
                                                    & \textbf{73.2 $\pm$ 0.5\%}
                                                    & \textbf{79.6 $\pm$ 0.2\%} \\ \hline
    \hline
\end{tabularx}
\end{table*}

\begin{table*}[!th]
\centering \caption{Comparison of g-U-Nets with and without graph
connectivity augmentation in terms of node classification accuracy
on Cora, Citeseer, and Pubmed datasets. }
\label{table:gunet_vs_gunet_no_aug}
\begin{tabularx}{\textwidth}{  lx{3.5cm}   YYY }
    \hline
    \textbf{Models}       & \textbf{Cora} & \textbf{Citeseer} & \textbf{Pubmed} \\ \hline\hline
    g-U-Nets without augmentation       & 83.7 $\pm$ 0.7\%   & 72.5 $\pm$ 0.6\% & 79.0 $\pm$ 0.3\% \\ \hline
    \textbf{g-U-Nets (Ours)}                         & \textbf{84.4 $\pm$ 0.6\%}
                                                    & \textbf{73.2 $\pm$ 0.5\%}
                                                    & \textbf{79.6 $\pm$ 0.2\%} \\ \hline
    \hline
\end{tabularx}
\end{table*}

\begin{table*}[!th]
\centering \caption{Comparison of different network depths in terms
of node classification accuracy on Cora, Citeseer, and Pubmed
datasets. Based on g-U-Nets, we experiment with different network
depths in terms of the number of blocks in encoder and decoder
parts.} \label{table:depth}
\begin{tabularx}{\textwidth}{  YYYY }
    \hline
    \textbf{Depth}   & \textbf{Cora} & \textbf{Citeseer} & \textbf{Pubmed} \\ \hline\hline
    2                & 82.6 $\pm$ 0.6\%   & 71.8 $\pm$ 0.5\% & 79.1 $\pm$ 0.3\% \\ \hline
    3                & 83.8 $\pm$ 0.7\%   & 72.7 $\pm$ 0.7\% & 79.4 $\pm$ 0.4\% \\ \hline
    4                & \textbf{84.4 $\pm$ 0.6\%}
                     & \textbf{73.2 $\pm$ 0.5\%}
                     & \textbf{79.6 $\pm$ 0.2\%} \\ \hline
    5                & 84.1 $\pm$ 0.5\%   & 72.8 $\pm$ 0.6\% & 79.5 $\pm$ 0.3\% \\ \hline
    \hline
\end{tabularx}
\end{table*}

\begin{table*}[!th]
\centering \caption{Comparison of the g-U-Nets with and without
gPool or gUnpool layers in terms of the node classification accuracy
and the number of parameters on Cora dataset.} \label{table:param}
\begin{tabularx}{\textwidth}{  lx{3cm}   YYY}
    \hline
    \textbf{Models}       & \textbf{Accuracy} & \textbf{\#Params} & \textbf{Ratio of increase} \\ \hline\hline
    g-U-Nets without gPool or gUnpool        & 82.1 $\pm$ 0.6\%           & 75,643 & 0.00\% \\ \hline
    \textbf{g-U-Nets (Ours)} & \textbf{84.4 $\pm$ 0.6\%}  & 75,737 & 0.12\% \\ \hline
    \hline
\end{tabularx}
\end{table*}

\subsection{Graph Connectivity Augmentation Study}

In the above experiments, we employ gPool layers with graph connectivity
augmentation by using the $2^{nd}$ graph power in Section~\ref{sec:aug}. Here,
we conduct experiments on node classification tasks to investigate the
benefits of graph connectivity augmentation based on g-U-Nets. We remove the
graph connectivity augmentation from gPool layers while keeping other settings
the same for fairness of comparisons. Table~\ref{table:gunet_vs_gunet_no_aug}
provides comparison results between g-U-Nets with and without graph
connectivity augmentation. The results show that the absence of graph
connectivity augmentation will cause consistent performance degradation on all
of three datasets. This demonstrates that graph connectivity augmentation via
$2^{nd}$ graph power can help with the graph connectivity and information transfer
among nodes in sampled graphs.


\subsection{Network Depth Study of Graph U-Nets}\label{sec:exp_depth}

Since the network depth in terms of the number of blocks in encoder
and decoder parts is an important hyper-parameter in the g-U-Nets,
we conduct experiments to investigate the relationship between
network depth and performance in terms of node classification
accuracy. We use different network depths on node classification
tasks and report the classification accuracies. The results are
summarized in Table~\ref{table:depth}. We can observe from the
results that the performance improves as network goes deeper until a
depth of 4. The over-fitting problem happens in deeper networks and
prevents networks from improving when the depth goes beyond that. In
image segmentation, U-Net models with depth 3 or 4 are commonly
used~\citep{badrinarayanan2017segnet,cciccek20163d}, which is
consistent with our choice in experiments. This indicates the
capacity of gPool and gUnpool layers in receptive field enlargement
and high-level feature encoding even working with very shallow
networks.


\subsection{Parameter Study of Graph Pooling Layers}\label{sec:exp_param}

Since our proposed gPool layer involves extra parameters, we compute
the number of additional parameters based on our g-U-Nets. The
comparison results between g-U-Nets with and without gPool or
gUnpool layers on dataset Cora are summarized in
Table~\ref{table:param}. From the results, we can observe that gPool
layers in U-Net model only adds 0.12\% additional parameters but can
promote the performance by a margin of 2.3\%. We believe this
negligible increase of extra parameters will not increase the risk
of over-fitting. Compared to g-U-Nets without gPool or gUnpool
layers, the encoder-decoder architecture with our gPool and gUnpool
layers yields significant performance improvement.

\section{Conclusion}

In this work, we propose novel gPool and gUnpool layers in g-U-Nets
networks for network embedding. The gPool layer implements the
regular global $k$-max pooling operation on graph data. It samples a
subset of important nodes to enable high-level feature encoding and
receptive field enlargement. By employing a trainable projection
vector, gPool layers sample nodes based on their scalar projection
values. Furthermore, we propose the gUnpool layer which applies
unpooling operations on graph data. By using the position
information of nodes in the original graph, gUnpool layer performs
the inverse operation of the corresponding gPool layer and restores
the original graph structure. Based on our gPool and gUnpool layers,
we propose the graph U-Nets~(g-U-Nets) architecture which uses a
similar encoder-decoder architecture as regular U-Net on image data.
Experimental results demonstrate that our g-U-Nets achieve
performance improvements as compared to other GNNs on transductive
learning tasks. To avoid the isolated node problem that may exist in
sampled graphs, we employ the $2^{nd}$ graph power to improve graph
connectivity. Ablation studies indicate the contributions of our
graph connectivity augmentation approach.

\clearpage

\section*{Acknowledgments}

This work was supported in part by National Science Foundation grants
IIS-1908166 and IIS-1908198.


\bibliography{deep}
\bibliographystyle{icml2019}

\end{document}